\pdfoutput=1

\documentclass[11pt]{article}

\usepackage[]{acl}

\usepackage{times}
\usepackage{latexsym}
\usepackage{authblk}
\usepackage{adjustbox}
\usepackage[T1]{fontenc}

\usepackage[utf8]{inputenc}

\usepackage{microtype}

\usepackage{inconsolata}
\usepackage{multirow}
\usepackage{todonotes}
\usepackage{xcolor}
\usepackage{xspace}
\usepackage[inline]{enumitem}
\usepackage{tikz}
\usepackage{xcolor}
\usepackage{hyperref}
\usepackage{tcolorbox}
\usetikzlibrary{shapes.geometric}
\newcommand*\circled[1]{\tikz[baseline=(char.base)]{
            \node[shape=circle,draw,inner sep=2pt] (char) {#1};}}

\newcommand{\orig}{{\textsc{Orig}}\xspace}
\newcommand{\btrain}{{\textsc{B-Train}}\xspace}
\newcommand{\ttrain}{{\textsc{T-Train}}\xspace}
\newcommand{\zs}{{\textsc{ZS}}\xspace}
\newcommand{\ttest}{{\textsc{TT}}\xspace}
\newcommand{\te}{{\textsc{TE}}\xspace}
\newcommand{\bt}{{\textsc{BT}}\xspace}

\title{Translation Errors Significantly Impact \\ Low-Resource Languages in Cross-Lingual Learning}

\newcommand*{\emails}{%
   \normalsize \texttt{\{ashishagrawal,barah,pjyothi\}@cse.iitb.ac.in}
}

\author{\textbf{Ashish Sunil Agrawal$^*$}}
\author{\textbf{Barah Fazili$^*$}}
\author{\textbf{Preethi Jyothi}}
\affil{Indian Institute of Technology Bombay, Mumbai, India}
\affil[ ]{ \vspace{-0.25cm} }
\affil[ ]{ \emails }

\begin{document}
\maketitle
\def\thefootnote{*}\footnotetext{These authors contributed equally to this work.}\def\thefootnote{\arabic{footnote}}

\begin{abstract}
Popular benchmarks (e.g., XNLI) used to evaluate cross-lingual language understanding consist of parallel versions of English evaluation sets in multiple target languages created with the help of professional translators. When creating such parallel data, it is critical to ensure high-quality translations for all target languages for an accurate characterization of cross-lingual transfer. In this work, we find that translation inconsistencies \emph{do exist} and interestingly they \emph{disproportionally impact low-resource languages} in XNLI. To identify such inconsistencies, we propose measuring the gap in performance between zero-shot evaluations on the human-translated and machine-translated target text across multiple target languages; relatively large gaps are indicative of translation errors. We also corroborate that translation errors exist for two target languages, namely Hindi and Urdu, by doing a manual reannotation of human-translated test instances in these two languages and finding poor agreement with the original English labels these instances were supposed to inherit.\footnote{Our code is available at  \href{https://github.com/csalt-research/translation-errors-crosslingual-learning}{https://github.com/translation-errors}}
\end{abstract}

\section{Introduction}

Multilingual benchmarks, such as XNLI, XTREME, play a vital role in assessing the cross-lingual generalization of multilingual pretrained models \citep{conneau-etal-2018-xnli, hu2020xtreme}. Typically, these benchmarks involve translating development and test sets from English into different target languages using professional human translators.
However, such a translation process is susceptible to human errors and could lead to incorrect estimates of cross-lingual transfer to target languages. We find translation errors do emerge and they disproportionately affect translations in certain low-resource languages such as Hindi and Urdu.%
\footnote{In the context of multilingual models, we refer to a language as low (or high)-resource based on the proportion of its data used in model pretraining. XLMR~\citep{conneau-etal-2020-unsupervised} is pretrained on the CC-100 corpus that includes roughly 50GB each of data from \emph{high-resource} languages such as French, Greek and Bulgarian, and only 20.2GB, 5.7GB and 1.6GB of data in \emph{low-resource} languages such as Hindi, Urdu and Swahili, respectively.}
%
%
%
\begin{figure}[t!]
    \centering    \includegraphics[width=0.99\linewidth]{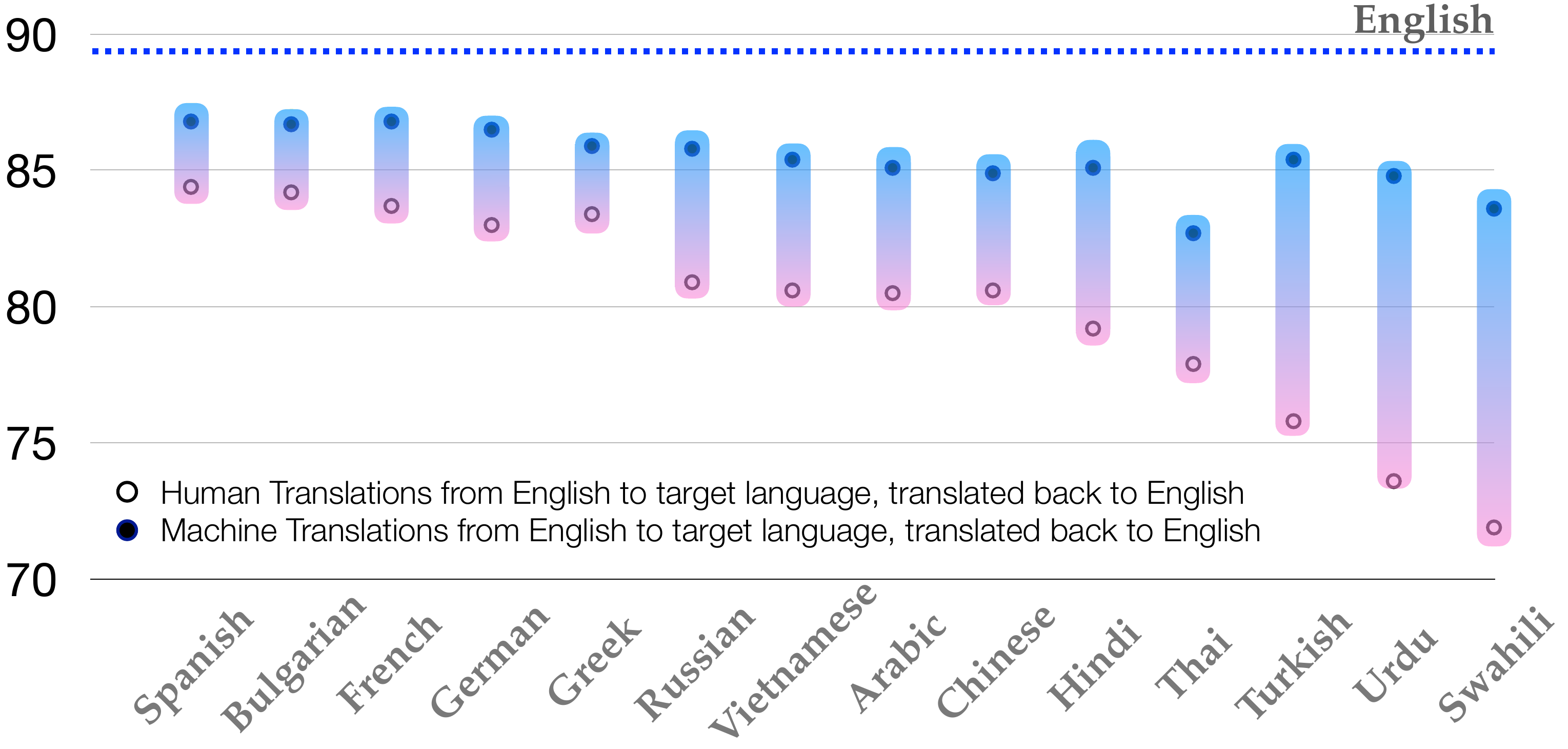}
\caption{XNLI performance gap by evaluating on translations of human-annotated data in target languages versus paraphrases of the original English data via backtranslations pivoted on each target language.}
   \label{fig:intro}
\end{figure}
%


Consider the well-known Cross-Lingual Natural Language Inference (XNLI) benchmark~\cite{conneau-etal-2018-xnli} that contains human translations of English premise-hypothesis pairs (with the labels reproduced from English) into 14 typologically-diverse target languages. Prior work raised concerns about whether the semantic relationships between premise and hypothesis are preserved in such human translations, but did not probe into this issue further~\cite{artetxe-etal-2020-translation,artetxe2023revisiting}. We find that there are indeed errors introduced in the human translations leading to label inconsistencies and that this issue disproportionately affects low-resource languages. 

To visualize the impact of low-quality translations on low-resource languages, Figure~\ref{fig:intro} compares zero-shot XNLI performance on all 14 target languages using the XLMR model~\citep{conneau-etal-2020-unsupervised} finetuned on English NLI with the following two input types:
\begin{enumerate*}
\item Human translations of the original English NLI instances to the target language from XNLI, translated back to English.
\item Machine translations of the original English NLI instances to the target language, translated back to English.
\end{enumerate*}
We see a clear differential trend with larger gaps between the (scores over the) two input types for low-resource languages such as Swahili, Urdu and Turkish (appearing on the right) and smaller gaps for high-resource languages such as Spanish, German and French (appearing on the left). We also observe that the \emph{cross-lingual transfer gap} when comparing the performance of human-translations for each target language with that of English (the latter shown as a dotted line) is largely overestimated for low-resource languages. 

To summarize, our main contributions are:
\begin{enumerate}[label=\protect\circled{\arabic*}]
\item We highlight the problem of translation errors in XNLI disproportionately affecting low-resource languages, and propose a practical way of identifying low-quality human translations by comparing their performance with machine translations derived from the original English sentences. 
\item We find that the translation errors persist under various train/test settings, including training data derived from machine-translations and paraphrases via backtranslations.
\item For two low-resource languages Hindi and Urdu, we manually annotate a subset of NLI data and find large discrepancies in the newly annotated labels when compared to the labels projected from the original English sentences. 

\end{enumerate}

\begin{table*}[t!]
\centering
\begin{tabular}{| p{0.04\linewidth}|p{0.02\linewidth} p{0.02\linewidth}p{0.02\linewidth}p{0.02\linewidth}p{0.02\linewidth} p{0.02\linewidth}p{0.02\linewidth}p{0.02\linewidth}p{0.02\linewidth}p{0.02\linewidth}p{0.02\linewidth}p{0.02\linewidth}p{0.02\linewidth}p{0.02\linewidth}p{0.02\linewidth} p{0.03\linewidth}|}
\hline 
\textbf{\small test} & \textbf{\small en} & \textbf{\small fr} &  \textbf{\small es} & \textbf{\small de} & \textbf{\small el} & \textbf{\small bg} & \textbf{\small ru} & \textbf{\small tr} & \textbf{\small ar} & \textbf{\small vi} & \textbf{\small th} & \textbf{\small zh} & \textbf{\small hi} & \textbf{\small sw} & \textbf{\small ur} & \textbf{\small avg}\\
\hline
\multirow{1}{*}{\small ZS} 
  &  \small 89.3 &  \small 83.5 &  \small 84.8 &  \small 83.4 &  \small 82.4 & \small 83.7 & \small 80.5 & \small 79.4 & \small 79.2 & \small 79.9 &  \small 78.3 &  \small 79.4 & \small 77.2 & \small 72.7 & \small 74.0 & \small 79.9\\
\hline
\multirow{1}{*}{\small TT-g} 
  & - &  \small 83.7 & \small 84.4 & \small 83.0 & \small 83.4 & \small 84.2 & \small 80.9 & \small 75.8 & \small 80.5 & \small 80.6 & \small 77.9 &  \small 80.6 &  \small 79.2 & \small 71.9 & \small 73.6 & \small 79.9\\
\hline
\hline
\multirow{1}{*}{\small TE-g} 
  & - &  \small \underline{85.3} &  \small \underline{85.9} &  \small \underline{85.9} &  \small \underline{84.8} &  \small \underline{86.1} &  \small \underline{84.9} &  \small \underline{83.8} &  \small \underline{82.7} &  \small \underline{84.0} &  \small \underline{82.0} & \small \underline{84.3}  &  \small \underline{82.1} &  \small \underline{77.3} &  \small \underline{81.8} & \small \underline{83.6}\\
\hline
\multirow{1}{*}{\small BT-g} 
   & - &  \small \textbf{86.6} &  \small \textbf{86.8} &  \small \textbf{86.5} &  \small \textbf{85.9} &  \small \textbf{86.7} &  \small \textbf{85.8} &  \small \textbf{85.4} &  \small \textbf{85.1} &  \small \textbf{85.4} &  \small \textbf{82.7} &  \small \textbf{84.9} &  \small \textbf{85.1} &  \small \textbf{83.6} &  \small \textbf{84.8} & \small \textbf{85.4}\\
\hline
\multirow{1}{*}{\small $\Delta$-g} 
   & & \small{2.9}  & \small{2} & \small{3.1} & \small{2.5} & \small{2.5}  & \small{4.9} & \small{6} & \small{4.6} & \small{4.8} & \small{4.4} & \small{4.3} & \small{5.9}  & \small{10.9} & \small{10.8}  & \small{4.9} \\
\hline
\end{tabular}
\caption{\label{orig}
Results of \orig (model trained on original English data) evaluated on different test set variants described in Section~\ref{sec:datasettings}. -g refers to using Google-translate as the translator. Highest scores in each column are shown in bold and next highest is underlined.
}
\end{table*}

\begin{table*}[t!]
\centering
\begin{tabular}{| p{0.04\linewidth}|p{0.02\linewidth} p{0.02\linewidth}p{0.02\linewidth}p{0.02\linewidth}p{0.02\linewidth} p{0.02\linewidth}p{0.02\linewidth}p{0.02\linewidth}p{0.02\linewidth}p{0.02\linewidth}p{0.02\linewidth}p{0.02\linewidth}p{0.02\linewidth}p{0.02\linewidth}p{0.02\linewidth} p{0.03\linewidth}|}
\hline
\textbf{\small test} &  \textbf{\small en} & \textbf{\small fr} &  \textbf{\small es} & \textbf{\small de} & \textbf{\small el} & \textbf{\small bg} & \textbf{\small ru} & \textbf{\small tr} & \textbf{\small ar} & \textbf{\small vi} & \textbf{\small th} & \textbf{\small zh} & \textbf{\small hi} & \textbf{\small sw} & \textbf{\small ur} & \textbf{\small avg}\\
\hline
\multirow{1}{*}{\small ZS} 
&  \small 89.2 &  \small 84.5 & \small 85.9 & \small 84.6 & \small 84.3 & \small 85.5 & \small 82.9 & \small 81.0 & \small 81.8 & \small 82.6 & \small 79.8 &  \small 80.9 &  \small 79.6 & \small 74.7 & \small 75.6 & \small 81.7\\
\hline
\multirow{1}{*}{\small TT-g} 
  & - & \small 84.8 & \small 86.5 & \small 84.1 & \small 85.1 & \small 85.9 & \small 82.7 & \small 78.9 & \small 83.1 & \small 82.7 & \small 80.4 & \small 82.6 & \small 81.4 & \small 74.9 & \small 76.9 & \small 82.1\\
\hline
\hline
\multirow{1}{*}{\small TE-g} 
  & - & \small \underline{86.6} & \small \underline{87.0} & \small \underline{86.9} & \small \underline{85.5} & \small \underline{86.4} & \small \underline{86.4} & \small \underline{84.3} & \small \underline{84.6} & \small \underline{84.9} & \small \underline{83.3} & \small \underline{84.6} & \small \underline{83.5} & \small \underline{78.9} & \small \underline{82.9} & \small \underline{84.7}\\
\hline
\multirow{1}{*}{\small BT-g} 
  & - & \small \textbf{88.0} & \small \textbf{87.7} & \small \textbf{87.6} & \small \textbf{86.7} & \small \textbf{87.5} & \small \textbf{87.1} & \small \textbf{85.9} & \small \textbf{86.4} & \small \textbf{86.2} & \small \textbf{84.2} & \small \textbf{85.9} & \small \textbf{85.9} & \small \textbf{85.4} & \small \textbf{86.1} & \small \textbf{86.5}\\
\hline
\multirow{1}{*}{\small $\Delta$-g} 
   & & \small{3.2}  & \small{1.2} & \small{2.5} & \small{1.6} & \small{1.6}  & \small{4.2} & \small{4.9} & \small{3.3} & \small{3.5} & \small{3.8} & \small{3.3} & \small{4.5}  & \small{10.5} & \small{9.2}  & \small{4.3} \\
\hline
\end{tabular}
\caption{\label{btrain}
Results of \btrain on different test set variants described in Section~\ref{sec:datasettings}. -g refers to using Google-translate as the translator.}
\end{table*}

\section{Experimental Setup}


\subsection{Tasks and Models}
\label{sec:tasks}
%
Our main focus is on the popular XNLI~\cite{conneau-etal-2018-xnli} benchmark, which is a three-way classification task to check whether a premise entails, contradicts or is neutral to a hypothesis. Parallel to English NLI \citep{bowman-etal-2015-large, williams-etal-2018-broad}, XNLI consists of development sets (2490 instances) and test sets (5010 instances) in 14 typologically-diverse languages\footnote{ Languages include French (fr), Spanish (es), German (de), Greek (el), Bulgarian (bg), Russian (ru), Turkish (tr), Arabic (ar), Vietnamese (vi), Thai (th), Chinese (zh), Hindi (hi), Swahili (sw) and Urdu (ur).} Translation-based gap analysis on two other multilingual tasks (MLQA  and PAWSX) is included in Appendix~\ref{more_tasks}.

We use XLM-Roberta (XLMR)~\cite{conneau-etal-2020-unsupervised} as the pretrained multilingual model in all our experiments. (Appendix~\ref{app:bert} reports scores using mBERT~\cite{devlin2019bert} for XNLI that follow the same trends.) 

\subsection{Training and Test Variants \label{sec:datasettings}}

\citep{artetxe-etal-2020-translation} showed that using machine-translated data to finetune the pretrained model helps it generalize better to both machine and human-translated test data. Motivated by this finding, we construct the following training variants:
\begin{enumerate}[label=\protect\circled{\arabic*}]
    \item \orig: Original English training data.
    \item Backtranslated-train (\btrain): English paraphrases of the original English data via backtranslations, with Spanish as a pivot. 
\end{enumerate}
\btrain is a training variant introduced in~\citep{artetxe-etal-2020-translation} that we adopt in our work. \\

\noindent We also evaluate on the following four variants of test data:
\begin{enumerate}[label=\protect\circled{\arabic*}]
    \item Zero-shot (\zs): Human-translated dev/test sets in the target languages. 
    \item Translate-test (\ttest): Machine translations of target language dev/test sets to English.
    \item Translate-from-English (\te): Machine translations of original English to the target languages.
    \item Backtranslation-via-target (\bt): Machine translations of original English to the target language and back to English.  
\end{enumerate}

We use two translation systems to create the above variants: 1) A state-of-the-art open-source multilingual translation model from the No Language Left Behind (NLLB) project~\cite{https://doi.org/10.48550/arxiv.2207.04672}, and 2) Google's Cloud Translate API.%
\footnote{https://cloud.google.com/translate}
%
Due to the prohibitive cost of the latter for the creation of training data, we use NLLB to create all our training variants (unless specified otherwise).%
\footnote{We found NLLB to be poor in quality when translating from English to Chinese. We used the M2M translation system~\citep{fan2020englishcentric} for English-to-Chinese that was far superior.}
Test variants were created using both translation systems. More implementation details and translation-related details are provided in Appendix~\ref{appntraining} and Appendix~\ref{appntranslation}. Some of the types of translation errors in the human-translated dev/test sets in \zs and \ttest are illustrated in Appendix~\ref{error_analysis}.

\section{Cross-lingual Transfer Gap in XNLI}
\label{main_results}
\subsection{Using Original English NLI Train Set}
Table~\ref{orig} presents XNLI accuracy scores for all four test variants using \orig training data. Test translations are generated using both NLLB (-n) and Google Translate (-g) (Numbers for NLLB translations are present in Appendix \ref{app:nllb}). $\Delta$-g in Table~\ref{orig} refers to the performance gap when using human vs. machine translations. It is the difference between the accuracy for \bt-g (machine-translated target language text) and the best accuracy among \zs and \ttest-g (human-translated target language text). It is striking that $\Delta$-g values for low-resource languages like Urdu and Swahili are as high as $10.8$ and $10.9$, respectively, and as low as $2.9$ and $2$ for high-resource languages like French and Spanish, respectively.


\subsection{Using Translated Train Sets}
\label{sec:ttrain}

Table~\ref{btrain} shows test accuracies using an XLMR model finetuned on \btrain. Across all target languages and all test set variants, we see consistent improvements in performance compared to \orig in Table~\ref{orig}. This is consistent with the observation in~\citep{artetxe-etal-2020-translation} that finetuning on backtranslation-driven paraphrases helps generalize better to both human and machine translated test sets. Interestingly, even with the overall improvements using \btrain, the large performance gap between \zs and \te (and \ttest and \bt) for low-resource languages like Urdu and Swahili persists.\footnote{We ran a Wilcoxon signed-rank test comparing accuracies from the ORIG model between the ZS test sets and BT-g test sets across all 14 languages. Performance on BT-g is significantly better (at p < 0.001) than on ZS test sets. We similarly found that the accuracies from the superior B-Train model is also significantly better (at p < 0.001) on the BT-g test sets compared to the ZS test sets.} 

\paragraph{Overestimated Cross-lingual Gap.} 
Based on ~\citet{hu2020xtreme}, we compute cross-lingual transfer gap as the difference between English accuracy and the average of accuracy scores across all other languages. From Table~\ref{btrain}, the previously reported cross-lingual gap was 7 using ZS, which reduces to 2.7 using \bt-g. The largest gaps for an individual language were previously 14.5 and 13.6 for Swahili and Urdu (the delta of their zero-shot scores wrt English test set scores) and have now reduced to 3.8 and 3.1 with \bt-g, respectively. This suggests a quick recipe for a quality check of human translations. For target languages supported by machine-translation systems, the performance gap between either \zs and \te or between \ttest and \bt could be a quick way to check whether the human translations might have issues during the data collection phase (thus yielding large  gap values). 


\section{Human Evaluation}
\label{sec:humaneval}

For two low-resource languages Hindi and Urdu, we reannotate a subset of the human-translations with NLI labels and check how well they match the labels inherited from the original English text. We pick random, non-overlapping sets of 200 instances each in English, Hindi and Urdu and get them relabelled by native speakers. (Appendix \ref{annotationdetails} provides more annotation details.) The new labels matched the original labels $90.5\%$, $66.5\%$ and $60\%$ of the time for English, Hindi and Urdu, respectively. This clearly highlights the large drop in label agreement for Hindi and Urdu compared to English, with relative reductions of 24\% and 30.5\% for Hindi and Urdu, respectively. In~\citet{conneau-etal-2018-xnli}, the same experiment was conducted using English and French and the original labels were recovered $85\%$ and $83\%$ of the time, respectively. The authors concluded there was no loss of information in the translations. However, we find there to be a significant loss of information in translations for languages such as Hindi and Urdu.

To verify if machine translations (\te) (rather than XNLI's human translations (\orig)) align better with the labels from the original English, we relabel 200  instances translated from English to Hindi and Urdu (via Google Translate). The annotators recovered the ground-truth labels $80\%$ and $71\%$ of the time for Hindi and Urdu, respectively, highlighting that label inconsistencies in Hindi/Urdu human translations (\orig) are significantly worse than with machine translations (\te). 
%

\section{Attention-based Analysis} 
We assess how the attention distributions learned for XNLI over the English test instances correlate with the attention distributions learned for human-annotated Hindi/Urdu/Swahili test instances and Google-translated (English to) Hindi/Urdu/Swahili test instances. 
For each correctly predicted English instance, we consider both human-translated (HT) and machine-translated (MT) target language translations and compute word alignments between English and these translations using awesome-align~\cite{dou2021word}. Aligned words whose attention score is greater than the mean attention score for the sequence are counted and normalized by the total number of such words in a sequence. Finally, we compute an average over all these overlap fractions across instances in the dataset. These mean overlap scores shown in Table~\ref{attention-scores} are computed separately using the human translations (HT) and machine translations (MT). For all three languages, we find the overlap fraction to be higher for the Google-translated sentences compared to the human-translated sentences. This suggests that MT aligns better with the original English text compared to HT. 

Since MT is typically more literal than human translations, thus resulting in more one-to-one aligned word pairs across the MT translations, it is not entirely surprising that we would see larger overlap fractions using MT translations in Table 3. We were also interested in the gap between the overlap fractions across MT and human translations across different languages. We observe that the gap between human and MT translations in terms of the overlap fractions is smaller for a high-resource language like French $(1.7\%)$, as opposed to Urdu $(5.3\%)$, Hindi $(2.8\%)$ or Swahili $(2.6\%)$.
\begin{table}[t!]
\centering
\begin{tabular}{|p{0.16\linewidth}|p{0.14\linewidth}|p{0.14\linewidth}|p{0.14\linewidth}|p{0.14\linewidth}|}
\hline
 \small\textbf{text/lang} & \small\textbf{ur} & \small \textbf{hi} & \small \textbf{sw} & \small \textbf{fr}\\
 \hline
\small HT & \small{0.375} & \small {0.392}  & \small {0.396} & \small{0.594}\\
\hline
\small MT & \small{\textbf{0.428}} & \small \textbf{0.42}  & \small \textbf{0.422} & \small \textbf{0.611}\\

\hline
\end{tabular}
\caption{\label{attention-scores}Aggregate attention scores over aligned words in Human Translated (HT) and Machine Translated (MT) XNLI test instances with parallel English data.
}
\end{table}
%
\begin{table*}[t!]
\centering
\begin{adjustbox}{width=\textwidth}
\begin{tabular}{p{0.12\linewidth}p{0.12\linewidth}p{0.12\linewidth}p{0.13\linewidth}p{0.08\linewidth}p{0.3\linewidth}}
\hline
\textbf{\small Premise} & \textbf{\small Hypothesis} & \textbf{\small En-Premise} & \textbf{\small En-Hypothesis} & \textbf{\small Label/Pred} & \textbf{\small Comment}\\
\hline
\small{Aise hi choti si baatein bhane mera karm par ek bada antar bana diya} & \small Mei kuch hasil karne ki koshish kar raha tha. & \small Little things like that made a big difference in what I was trying to do. & \small I was trying to accomplish something. & \small E/N & \small Incorrect translation of premise changes the relationship between the label and the premise-hypothesis pair.\\
\hline
\small Mei tumhe ek ghante mei wapas phone karta hoo, ve kehte hai. & \small Usne kaha ki ve bol rahe the. & \small I'll call you back in about an hour, he says. & \small He said they were done speaking. & \small C/E & \small Hypothesis is incorrectly translated leading to a change in meaning (i.e "they were done speaking" is translated to "they were speaking").\\
\hline

\small Wo qaed nahin rehna chahte they  & \small Unhe kuch mawaqe par pakda ja sakta tha lekin wo is se bachna chahte they & \small They didn't want to stay captive.& \small They had been captured at some point but wanted to escape. & \small N/C & \small Tense is incorrect in the translation of the hypothesis. The premise implies that they have already been captured while the incorrect translation implies that they did not want to get caught, hence predicting a contradiction. \\
\hline
\small Ye tha, ye ek khoobsoorat din tha & \small Aj ek aramdah din tha & \small That was, that was a pretty scary day. & \small It was a relaxing day. & \small C/N & \small Tense is incorrectly altered to present and "pretty scary" is translated to simply "khoobsoorat"(pretty), thus inverting the overall sentiment.\\
\hline
\end{tabular}
\end{adjustbox}
\caption{\label{hiexamples}
Semantically incorrect examples of premise-hypothesis pairs in Hindi (first two) and Urdu (latter two). E, N and C implies entailment, neutral and contradiction labels.}
\end{table*}
\section{Impact of Using Translations for Multilingual Datasets}
\label{error_analysis}
Table~\ref{hiexamples} highlights a few examples of premise-hypothesis pairs in XNLI's Hindi and Urdu that are no longer semantically consistent with the original labels (copied from English) after translation. These examples would be flagged as having prediction errors when in fact the predictions are reasonable given the semantic deviations in the human-translated Hindi/Urdu sentences from the original English sentences.

While Table~\ref{hiexamples} shows examples of errors, translation issues might not always be errors and could just be deviations due to unfamiliar phrases or English-specific nuances that do not get adequately captured in the translations. For example, we show a snippet of a premise below:\\
\textit{English premise}: “but no … is what you see down here so it's nice with me working at home because i can wear pants”\\
\textit{Google translated premise}: lekin nahi … jo ap yahan neeche dekh rahe hain isliye mere saath ghar par kaam karna accha hai kyonki main pants pehen sakti hun\\
\textit{Human translated premise}: lekin nahi …  jo ki ap neeche dekhte hi hain, isliye mere saath ghar par kaam karna accha hai kyonki main pants pehen sakti hun\\
The phrase "nice with me working at home" was incorrectly translated as "mere saath ghar par kaam karna," which back-translates to “work at home with me.” This misinterpretation may stem from the unfamiliar phrase in English.

As NLP systems improve, high-quality manual annotations are critical. With existing NLP systems already showing differential trends on high- versus low-resource languages~\cite{robinson-etal-2023-chatgpt}, it is increasingly important to insulate against translation inadequacies leading to label errors that predominantly affect low-resource languages. 

\section{Related Work}
There is growing interest in building multilingual benchmarks for the evaluation of cross-lingual transfer. E.g., XTREME~\cite{https://doi.org/10.48550/arxiv.1911.02116} covering a wide range of languages and tasks including XNLI~\cite{conneau-etal-2018-xnli}, XQuAD~\cite{xquad}, PAWS-X~\cite{paws-x} and MLQA~\cite{mlqa}. 
Recently, many extensions of XTREME: IndXTREME~\cite{indic-xtreme} focusing on 18 Indian languages, XTREME-R~\cite{ruder-etal-2021-xtreme} and XTREME-UP~\cite{ruder2023xtremeup} have also been released. 
Translation artifacts have only been studied in select prior works. \citep{10.5555/3013558.3013562} study how translations can alter sentiment labels in Arabic text. 
In very recent work, \citep{artetxe2023revisiting} advocate for the use of English-only finetuning using machine-translation systems. However, this relies on high-quality human translations in the target languages which we highlight needs to be carefully examined especially for low-resource languages.

\section{Conclusions}
This work studies the problem of translation irregularities in evaluation sets of multilingual benchmarks like XNLI that are created by translating English  into multiple target languages. We find that the translation sets of low-resource languages like Urdu, Swahili exhibit most inconsistencies while translations of high-resource languages like French, German are more immune to this problem. We suggest an effective way to check the quality of human translations by comparing performance with machine translations, and show how the cross-lingual transfer estimates can significantly vary with improved translations.

\section{Acknowledgements}
The last author would like to gratefully acknowledge a faculty grant from Google Research India supporting her research on multilingual models. The authors are also thankful to the anonymous reviewers for very constructive feedback.

\section{Limitations}
For tasks that have output labels directly corresponding to the input text (e.g., sequence labeling tasks like POS-tagging, question answering, etc.), it would be trickier to use our technique since  translations could change the word order and subsequently affect the output labels as well.   

We highlight the problem of the cross-lingual transfer gap for low-resource languages being mischaracterized due to poor performance on these languages stemming from poor-quality translations and not necessarily because the model has difficulty with the given target languages. We do not offer a solution to deal with translation errors. Rather, we ask for additional checks when collecting translations for low-resource languages. 

We identify that the existing translation datasets for low-resource languages in XNLI have inconsistencies. While we did not create manually-corrected versions of these translation sets, we will be releasing the machine-translated text from English to these target languages upon publication.

\section*{Ethics Statement}
We would like to emphasize our commitment to upholding ethical practices throughout this work. We aimed to ensure that human annotators received a fair compensation for their annotation efforts and was commensurate with the time and effort invested in their work. For translations using Google Translate, we used the paid Cloud API service in accordance with the terms and conditions of usage.

\bibliography{anthology,custom}

\appendix
\label{sec:appendix}

\section{Performance Gap Analysis for MLQA, PAWS-X}
\label{more_tasks}
Multilingual (Extractive) Question Answering (\citet{mlqa}, MLQA) consists of questions in English translated to six different languages including Arabic (ar), German (de), Spanish (es), Hindi (hi), Vietnamese (vi) and Chinese (zh) amounting to 5K instances in each target language. 
PAWS-X: A Cross-lingual Adversarial Dataset for Paraphrase Identification \cite{paws-x} consists of dev/test paraphrases in English translated to six different languages: French(fr), Spanish(es), German (de), Chinese (zh), Japanese (ja), and Korean (ko) with the help of human translators. 

\paragraph{MLQA.} For MLQA, we translate questions in the two low-resource languages, Hindi and Vietnamese, to English using NLLB (\ttest). We also create a \bt version of the original English questions (\ref{sec:datasettings}) using Hindi and Vietnamese as pivots. 

\begin{table}
\begin{tabular}
{|p{0.14\linewidth}|p{0.14\linewidth}p{0.14\linewidth}|p{0.14\linewidth}p{0.14\linewidth}|}
\hline
\textbf{\small F1/EM} & \textbf{\small en} & \textbf{\small hi} & \textbf{\small en} & \textbf{\small vi}\\
\small (\# sents) & \small (4918) & \small (4918 ) & \small (5495 ) & {\small (5495)}\\
\hline
\small \zs & \small 83.2/69.8 & \small 70.6/52.9 & \small 83.4/70.6 & \small 74.0/52.7\\
\small \ttest-n & \small - & \small 78.4/64.5 & \small -  & \small 74.9/61.3\\
\hline
\hline
\small \bt-n & \small - & \small \textbf{78.4}/\textbf{64.7} & \small - & \small \textbf{76.7}/\textbf{63.2}\\
\hline
\end{tabular}
\caption{\label{mlqares}
Results on \ttest-n and \bt-n MLQA test sets. \bt-n Hi indicates backtranslated data pivoted through Hindi, \ttest-n Hi indicates test set in Hi translated to En. (Note that for MLQA only questions are translated.)}
\end{table}

\begin{table}
\begin{tabular}
{|p{0.92\linewidth}|}
\hline
\textbf{\small Instructions}\\
\hline
\small Given premise and hypothesis, label each pair as "entailment", "contradiction" or "neutral" as follows: \\
\small 1. if hypothesis is entailed by the premise, it's an "entailment" ,\\
\small 2. if the hypothesis contradicts the premise (hypothesis cannot be True given the premise), it's a "contradiction",\\
\small 3. if the hypothesis is independent of the premise (hypothesis may or may not be True given the premise), it's a "neutral" relationship. \\
\hline
\end{tabular}
\caption{\label{instructions} Task description shared with the annotators for the NLI task
}
\end{table}

\begin{table*}[ht!]
\centering
\resizebox{1\linewidth}{!}{
\begin{tabular}{|ccccccccc|}
\hline
\small dev/test & \textbf{\small en} & \textbf{\small de} & \textbf{\small es} & \textbf{\small fr} & \textbf{\small ja} & \textbf{\small ko}& \textbf{\small zh} & \textbf{\small avg}\\
\small sents & \small (2000/2000) & \small (2000/2000) & \small (2000/2000) & \small (2000/2000) & \small (2000/2000) & \small (2000/2000) & \small (2000/2000) & - \\
\hline
\small \zs & \small 95/95.9 & \small 89/90.9 & \small 90.4/90.4 & \small 91.4/91.6 & \small 82.9/80.5 & \small 83.6/80.8 & \small 83.9/84.2 & \small 86.9/86.4\\
\small \ttest-n & \small - & \small 88.9/89.9   & \small 89.8/91  & \small 90.4/91.6  & \small 83/79 & \small 82.2/80.4   & \small  81.6/80.9  & \small 86.0/85.5\\
\hline
\hline
\small \te-n & \small - & \small 91.2/92.3  & \small 92.1/92.3 & \small 90.9/91.2  & \small 83.7/83.4  & \small 86.8/85.7  & \small 88.8/88.6  & \small 88.9/88.9\\
\small \bt-n & \small - & \small 90.6/91.5   & \small 91.6/92.2  & \small 90.8/90.8  & \small 81.9/80.6 & \small 84/84.4 & \small 89/88.2 & \small 88.0/88.0 \\
\hline
\end{tabular}
}
\caption{\label{paws-x}
Results on \zs, \te, \ttest, and \bt PAWS-X.}
\end{table*}

Table \ref{mlqares} shows TT and BT scores for Hindi are nearly identical and there is a small improvement using BT for Vietnamese compared to TT. This indicates that the professional annotators did not introduce semantic inconsistencies during translation for MLQA. In general, classification tasks like XNLI appear to be more susceptible to translation inconsistencies since the annotators are not made aware of the ground-truth labels during translation and are only asked to independently translate the premise/hypothesis pairs.

\paragraph{PAWS-X.} Table \ref{paws-x} shows the results of the different settings ZS, TE, TT, and BT for the six languages. The model used for inference is xlm-roberta-large trained on the English train set. TE is better than ZS mainly for Korean (by 4.9\% in test set) and Chinese (4.9\% in dev set) and is nearly equal for other languages. BT is better than TT again for Korean and Chinese and nearly equal for other languages. This indicates the presence of human translation inconsistency for the two languages.

\begin{table*}
\centering
\begin{tabular}{| p{0.065\linewidth}|p{0.03\linewidth} p{0.03\linewidth}p{0.03\linewidth}p{0.03\linewidth}p{0.03\linewidth} p{0.03\linewidth}p{0.03\linewidth}p{0.03\linewidth}p{0.03\linewidth}p{0.03\linewidth}p{0.03\linewidth}p{0.03\linewidth}p{0.03\linewidth}p{0.03\linewidth}p{0.03\linewidth}p{0.03\linewidth} |}
\hline
\textbf{dev} &  \textbf{en} & \textbf{fr} &  \textbf{es} & \textbf{de} & \textbf{el} & \textbf{bg} & \textbf{ru} & \textbf{tr} & \textbf{ar} & \textbf{vi} & \textbf{th} & \textbf{zh} & \textbf{hi} & \textbf{sw} & \textbf{ur} & \textbf{avg}\\
\hline
\multirow{1}{*}{\small XLMR} 
  &  \small 89.9 &  \small 84.2 &  \small 85.0 & \small 84.3 & \small 81.8 & \small 83.2 & \small 79.7 & \small 79.9 &  \small 79.2 &  \small 81.6 &  \small 78.0 &  \small 80.0 &  \small 78.3 &  \small 72.1 &  \small 74.6 & \small 80.8\\
\hline
\multirow{1}{*}{\small mBert} 
  &  \small 83.0 &  \small 74.9 &  \small 74.8 & \small 72.2 & \small 67.8 & \small 68.2 & \small 68.4 & \small 63.4 &  \small 65.4 &  \small 69.8 &  \small 54.8 &  \small 70.6 &  \small 61.5 &  \small 52.4 &  \small 53.3 & \small 66.7\\
\hline
\end{tabular}
\caption{\label{mbertxlmr}
Zero shot performance of \orig mBert and XLMR models on the XNLI target dev sets.
}
\end{table*}

\begin{table*}[t!]
\centering
\begin{tabular}{| p{0.04\linewidth}|p{0.02\linewidth} p{0.02\linewidth}p{0.02\linewidth}p{0.02\linewidth}p{0.02\linewidth} p{0.02\linewidth}p{0.02\linewidth}p{0.02\linewidth}p{0.02\linewidth}p{0.02\linewidth}p{0.02\linewidth}p{0.02\linewidth}p{0.02\linewidth}p{0.02\linewidth}p{0.02\linewidth} p{0.03\linewidth}|}
\hline 
\textbf{\small test} & \textbf{\small en} & \textbf{\small fr} &  \textbf{\small es} & \textbf{\small de} & \textbf{\small el} & \textbf{\small bg} & \textbf{\small ru} & \textbf{\small tr} & \textbf{\small ar} & \textbf{\small vi} & \textbf{\small th} & \textbf{\small zh} & \textbf{\small hi} & \textbf{\small sw} & \textbf{\small ur} & \textbf{\small avg}\\
\hline
\multirow{1}{*}{\small ZS} 
  &  \small 89.3 &  \small 83.5 &  \small 84.8 &  \small 83.4 &  \small 82.4 & \small 83.7 & \small 80.5 & \small 79.4 & \small 79.2 & \small 79.9 &  \small 78.3 &  \small 79.4 & \small 77.2 & \small 72.7 & \small 74.0 & \small 79.9\\
\hline
\multirow{1}{*}{\small TT-n} 
  & - &  \small 82.1 &  \small 83.1 & \small 80.7 & \small 82.3 & \small 82.6 & \small 79.3  & \small 75.9 &  \small 78.0 & \small 78.7  & \small 73.8  & \small 77.6 & \small 77.7  & \small 70.5 & \small 71.3 & \small 78.1\\
\hline
\hline
\multirow{1}{*}{\small BT-n} 
   & - &  \small \textbf{84.5} &  \small \underline{84.9} &  \small \underline{83.5} &  \small \underline{82.9} &  \small \underline{82.7} &  \small \underline{82.3} &  \small \underline{81.1} &  \small \underline{81.4} &  \small \textbf{82.4} &  \small \underline{76.4} & \small \underline{79.6} &  \small \textbf{82.9} &  \small \textbf{79.4} &  \small \textbf{80.8} & \small \underline{81.8}\\
\hline
\multirow{1}{*}{\small TE-n} 
  & - &  \small \underline{84.4} &  \small \textbf{85.5} &  \small \textbf{83.9} &  \small \textbf{83.6} &  \small \textbf{83.9} &  \small \textbf{83.4} &  \small \textbf{81.7} &  \small \textbf{81.5} &  \small \underline{81.9} &  \small \textbf{78.7} & \small \textbf{81.0}  &  \small \underline{82.1} &  \small \underline{77.0} &  \small \underline{80.3} & \small \textbf{82.1}\\
\hline
\multirow{1}{*}{\small $\Delta$-n} 
   & & \small{1}  & \small{0.7} & \small{0.5} & \small{1.2} & \small{0.2}  & \small{2.9} & \small{2.3} & \small{2.3} & \small{2.5} & \small{0.4} & \small{1.6} & \small{5.2}  & \small{6.7} & \small{6.8}  & \small{2.2} \\
\hline
\end{tabular}
\caption{\label{orig-nllb}
Results of \orig (model trained on original English data) evaluated on different test set variants described in Section~\ref{sec:datasettings}. -n refers to using NLLB as the translator. Highest scores in each column are shown in bold and next highest is underlined.
}
\end{table*}

\begin{table*}[t!]
\centering
\begin{tabular}{| p{0.04\linewidth}|p{0.02\linewidth} p{0.02\linewidth}p{0.02\linewidth}p{0.02\linewidth}p{0.02\linewidth} p{0.03\linewidth}p{0.02\linewidth}p{0.02\linewidth}p{0.02\linewidth}p{0.02\linewidth}p{0.02\linewidth}p{0.02\linewidth}p{0.02\linewidth}p{0.02\linewidth}p{0.02\linewidth} p{0.03\linewidth}|}
\hline
\textbf{\small test} &  \textbf{\small en} & \textbf{\small fr} &  \textbf{\small es} & \textbf{\small de} & \textbf{\small el} & \textbf{\small bg} & \textbf{\small ru} & \textbf{\small tr} & \textbf{\small ar} & \textbf{\small vi} & \textbf{\small th} & \textbf{\small zh} & \textbf{\small hi} & \textbf{\small sw} & \textbf{\small ur} & \textbf{\small avg}\\
\hline
\multirow{1}{*}{\small ZS} 
&  \small 89.2 &  \small 84.5 & \small 85.9 & \small 84.6 & \small 84.3 & \small \textbf{85.5} & \small 82.9 & \small 81.0 & \small 81.8 & \small 82.6 & \small 79.8 &  \small 80.9 &  \small 79.6 & \small 74.7 & \small 75.6 & \small 81.7\\
\hline
\multirow{1}{*}{\small TT-n} 
& - &  \small 84.0 & \small 85.7  &  \small 82.4 &  \small 84.4 & \small 84.4 & \small 81.8  & \small 78.9 & \small 81.0 & \small 80.9 & \small 77.4 & \small 80.5 & \small 80.5 & \small 73.6 & \small 74.4 & \small 80.7\\
\hline
\hline
\multirow{1}{*}{\small BT-n} 
& - &  \small \textbf{85.9} &  \small \textbf{86.8} &  \small \underline{85.1} &  \small \underline{84.8} &  \small 84.6 & \small \underline{84.3} &  \small \underline{82.8} &  \small \textbf{83.5} &  \small \textbf{84.2} &  \small \underline{79.3} & \small \underline{81.4} &  \small \textbf{84.8} & \small \textbf{81.9} & \small \textbf{82.5} & \small \textbf{83.7}\\
\hline
\multirow{1}{*}{\small TE-n} 
& - &  \small \underline{85.8} &  \small \underline{86.8} &  \small \textbf{85.2} &  \small \textbf{84.9} &  \small \underline{85.2} &  \small \textbf{84.6} &  \small \textbf{83.0} &  \small \underline{83.5} &  \small \underline{83.6} &  \small \textbf{80.6} &  \small \textbf{82.0} &  \small \underline{83.4} &  \small \underline{79.1} &  \small \underline{81.4} & \small \underline{83.5}\\
\hline
\multirow{1}{*}{\small $\Delta$-n} 
   & & \small{1.4}  & \small{0.9} & \small{0.6} & \small{0.5} & \small{-0.3} & \small{1.7} & \small{2} & \small{1.7} & \small{1.6} & \small{1.6} & \small{1.1} & \small{4.3}  & \small{7.2} & \small{6.9}  & \small{2} \\
\hline
\end{tabular}
\caption{\label{btrain-nllb}
Results of \btrain on different test set variants described in Section~\ref{sec:datasettings}. -n refers to using NLLB as the translator.}
\end{table*}

\begin{table*}[t!]
\centering
\begin{tabular}{| p{0.065\linewidth}|p{0.03\linewidth} p{0.03\linewidth}p{0.03\linewidth}p{0.03\linewidth}p{0.03\linewidth} p{0.03\linewidth}p{0.03\linewidth}p{0.03\linewidth}p{0.03\linewidth}p{0.03\linewidth}p{0.03\linewidth}p{0.03\linewidth}p{0.03\linewidth}p{0.03\linewidth}p{0.03\linewidth}p{0.03\linewidth} |}
\hline
\textbf{test} &  \textbf{en} & \textbf{fr} &  \textbf{es} & \textbf{de} & \textbf{el} & \textbf{bg} & \textbf{ru} & \textbf{tr} & \textbf{ar} & \textbf{vi} & \textbf{th} & \textbf{zh} & \textbf{hi} & \textbf{sw} & \textbf{ur} & \textbf{avg}\\
\hline
\multirow{1}{*}{\small ZS} 
&  \small 88.9 &  \small 84.8 &  \small 85.7 &  \small 84.8 &  \small 84.4 &  \small 85.0 &  \small 82.9 &  \small 80.9 &  \small 81.2 &  \small 81.9 & \small 78.9 &  \small 80.7 &  \small 79.6 & \small 74.9 & \small 75.9 & \small 81.7\\
\hline
\multirow{1}{*}{\small TT-n} 
& - &  \small 83.2 &  \small 84.5 & \small 82.4 & \small 83.9 & \small 84.1 & \small 81.3 & \small 78.4 & \small 80.6 & \small 80.7 &  \small 76.6 &  \small 79.7 & \small 80.1 & \small 73.1 & \small 74.2 & \small 80.2\\
\hline
\multirow{1}{*}{\small TT-g} 
& - &  \small 84.3  &  \small 85.9  &  \small 84.2 &  \small 84.8 &  \small 85.2 &  \small 82.8 &  \small 77.8 &  \small 82.5 &  \small 81.9 &  \small 79.9 &  \small 82.2 &  \small 81.1 &  \small 74.3 &  \small 76.0 & \small 81.6\\
\hline
\multirow{1}{*}{\small BT-n} 
& - &  \small 85.2 &  \small 86.2 &  \small 84.6 &  \small 84.8 &  \small 84.2 &  \small 83.9 &  \small 82.3 &  \small 83.3 &  \small 83.9 &  \small 79.2 &  \small 81.6 &  \small \underline{84.4} &  \small \underline{81.4} &  \small 81.9 & \small 83.4\\
\hline
\multirow{1}{*}{\small TE-n} 
& - &  \small 85.3 &  \small 86.3 &  \small 85.1 &  \small 84.4 &  \small 84.9 &  \small 84.7 &  \small 82.5 &  \small 83.1 &  \small 83.9 &  \small 79.9 &  \small 81.8 &  \small 83.0 &  \small 79.0 &  \small 81.4 & \small 83.2\\
\hline
\multirow{1}{*}{\small TE-g} 
& - &  \small \underline{86.2}  &  \small \underline{86.6}  & \small \underline{86.5}  &  \small \underline{85.1}  &  \small \underline{86.8}  &  \small \underline{86.0} &  \small \underline{83.9} &  \small \underline{84.1} &  \small \underline{85.0} &  \small \underline{82.7} &  \small \underline{84.5} &  \small 83.4 &  \small 79.4 &  \small \underline{82.8} & \small \underline{84.5}\\
\hline
\multirow{1}{*}{\small BT-g} 
& - &  \small \textbf{87.0} &  \small \textbf{87.3} &  \small \textbf{87.3} &  \small \textbf{86.7} &  \small \textbf{87.0} &  \small \textbf{86.7} &  \small \textbf{85.7} &  \small \textbf{86.0} &  \small \textbf{86.1} &  \small \textbf{83.8} &  \small \textbf{85.5} &  \small \textbf{85.8} &  \small \textbf{84.6} &  \small \textbf{85.5} & \small \textbf{86.1}\\
\hline
\multirow{1}{*}{\small $\Delta$-g} 
   & & \small{2.2}  & \small{1.4} & \small{2.5} & \small{1.9} & \small{1.8}  & \small{3.8} & \small{4.8} & \small{3.5} & \small{4.2} & \small{4.1} & \small{3.3} & \small{4.7}  & \small{9.7} & \small{9.5}  & \small{4.1} \\
\hline
\end{tabular}
\caption{\label{ttrain}
Results of \ttrain on different test set variants described in Section \ref{sec:datasettings}.}
\end{table*}

\begin{table*}
\centering
\begin{tabular}{| p{0.065\linewidth}|p{0.03\linewidth} p{0.03\linewidth}p{0.03\linewidth}p{0.03\linewidth}p{0.03\linewidth} p{0.03\linewidth}p{0.03\linewidth}p{0.03\linewidth}p{0.03\linewidth}p{0.03\linewidth}p{0.03\linewidth}p{0.03\linewidth}p{0.03\linewidth}p{0.03\linewidth}p{0.03\linewidth}p{0.03\linewidth} |}
\hline
\textbf{test} &  \textbf{en} & \textbf{fr} &  \textbf{es} & \textbf{de} & \textbf{el} & \textbf{bg} & \textbf{ru} & \textbf{tr} & \textbf{ar} & \textbf{vi} & \textbf{th} & \textbf{zh} & \textbf{hi} & \textbf{sw} & \textbf{ur} & \textbf{avg}\\
\hline
\multirow{1}{*}{\small \zs} 
&  \small 89.8 &  \small 85.1 &  \small 86.2 &  \small 84.6 &  \small 84.1 & \small 85.2 & \small 82.4 & \small 81.3 & \small 81.2 & \small 81.9 &  \small 79.3 &  \small 80.9 & \small 78.6 & \small 74.9 & \small 76.0 & \small 82.1\\
\hline
\multirow{1}{*}{\small \ttest-n} 
& - &  \small 84.2 &  \small 85.2 & \small 82.6 &  \small 84.8 & \small 84.8 & \small 81.9 &  \small 78.8 &  \small 81.7 & \small 81.1 & \small 78.2 & \small 80.3 & \small 80.7 & \small 73.8 &  \small 75.1 & \small 80.9\\
\hline
\multirow{1}{*}{\small \bt-n} 
& - &  \small \textbf{85.9} &  \small \textbf{86.6} &  \small \textbf{85.0} &  \small \textbf{85.0} & \small \textbf{85.2} &  \small \textbf{84.2} &  \small \textbf{83.2} &  \small \textbf{83.6} &  \small \textbf{84.8} &  \small \textbf{79.4} &  \small \textbf{81.9} &  \small \textbf{85.2} &  \small \textbf{82.1} & \small \textbf{82.8} & \small \textbf{83.9}\\
\hline
\multirow{1}{*}{\small \te-n} 
& - &  \small \textbf{85.9} &  \small \textbf{87.0} &  \small \textbf{85.2} &  \small \textbf{84.5} &  \small \textbf{85.3} &  \small \textbf{84.6} &  \small \textbf{83.1} &  \small \textbf{83.6} &  \small \textbf{84.2} &  \small \textbf{80.1} &  \small \textbf{82.7} &  \small \textbf{82.9} &  \small \textbf{78.7} &  \small \textbf{80.8} & \small \textbf{83.5}\\
\hline
\end{tabular}
\caption{\label{btesen}
Results of BT-enes (model trained on back-translated(en$\rightarrow$es$\rightarrow$ en) + original English train set) on different test set data settings \ref{sec:datasettings}, \textbf{-n} refers to using NLLB translator.
}
\end{table*}

\begin{table*}
\centering
\begin{tabular}{| p{0.065\linewidth}|p{0.03\linewidth} p{0.03\linewidth}p{0.03\linewidth}p{0.03\linewidth}p{0.03\linewidth} p{0.03\linewidth}p{0.03\linewidth}p{0.03\linewidth}p{0.03\linewidth}p{0.03\linewidth}p{0.03\linewidth}p{0.03\linewidth}p{0.03\linewidth}p{0.03\linewidth}p{0.03\linewidth}p{0.03\linewidth} |}
\hline
\textbf{test} &  \textbf{en} & \textbf{fr} &  \textbf{es} & \textbf{de} & \textbf{el} & \textbf{bg} & \textbf{ru} & \textbf{tr} & \textbf{ar} & \textbf{vi} & \textbf{th} & \textbf{zh} & \textbf{hi} & \textbf{sw} & \textbf{ur} & \textbf{avg}\\
\hline

\multirow{1}{*}{\small \zs} 
&  \small 87.4 & \small 82.9 & \small 84.2 & \small 82.7 & \small 83.4 & \small 83.4 & \small 81.1 & \small 80.8 & \small 79.9 & \small 80.4 & \small 78.1 &  \small 79.9 & \small 78.8 & \small 74.1 & \small 75.3 & \small 80.8\\
\hline
\multirow{1}{*}{\small \ttest-n} 
& - &  \small 81.7 & \small 82.6 & \small 80.1 & \small 82.2 & \small 82.3 & \small 80.3 & \small 76.2 & \small 79.4 & \small 79.3 & \small 75.8 &  \small 77.9 &  \small 78.5 & \small 72.2 & \small 72.5 & \small 78.6\\
\hline
\multirow{1}{*}{\small \bt-n} 
& - &  \small \textbf{83.9} & \small  \textbf{84.4} & \small  \textbf{83.4} &  \small \textbf{82.7} &  \small 81.8 &  \small \textbf{82.3} &  \small \textbf{80.1} &  \small \textbf{81.5} & \small  \textbf{82.2} & \small  \textbf{77.5} & \small \textbf{80.0} &  \small \textbf{83.3} & \small \textbf{79.9} & \small  \textbf{81.0} & \small \textbf{81.7}\\
\hline
\multirow{1}{*}{\small \te-n} 
& - &  \small \textbf{83.7} &  \small \textbf{84.9} &  \small \textbf{83.6} &  \small 83.0 & \small \textbf{83.5} &  \small \textbf{82.8} & \small  \textbf{81.5} &  \small \textbf{82.0} & \small  \textbf{82.3} &  \small \textbf{79.4} & \small \textbf{81.1} & \small  \textbf{82.7} & \small  \textbf{78.2} & \small \textbf{81.4} & \small \textbf{82.1}\\
\hline
\end{tabular}
\caption{\label{mthig}
Results of MT-hi-g (model trained on data translated to Hindi (en$\rightarrow$hi) using google-translate) on different test set data settings \ref{sec:datasettings}.
}
\end{table*}

\begin{table*}
\centering
\begin{tabular}{| p{0.065\linewidth}|p{0.03\linewidth} p{0.03\linewidth}p{0.03\linewidth}p{0.03\linewidth}p{0.03\linewidth} p{0.03\linewidth}p{0.03\linewidth}p{0.03\linewidth}p{0.03\linewidth}p{0.03\linewidth}p{0.03\linewidth}p{0.03\linewidth}p{0.03\linewidth}p{0.03\linewidth}p{0.03\linewidth}p{0.03\linewidth} |}
\hline
\textbf{test} &  \textbf{en} & \textbf{fr} &  \textbf{es} & \textbf{de} & \textbf{el} & \textbf{bg} & \textbf{ru} & \textbf{tr} & \textbf{ar} & \textbf{vi} & \textbf{th} & \textbf{zh} & \textbf{hi} & \textbf{sw} & \textbf{ur} & \textbf{avg}\\
\hline
\multirow{1}{*}{\small \zs} 
&  \small 87.2 &  \small 83.4 & \small 83.6 & \small 82.9 & \small 82.7 & \small 83.4 & \small 81.8 & \small 79.9 & \small 79.9 & \small 80.1 & \small 78.7 & \small 80.6 & \small 78.4 & \small 73.6 &  \small 74.9 & \small 80.7\\
\hline
\multirow{1}{*}{\small \ttest-n} 
& - & \small 82.2 & \small 83.6 & \small 80.6 & \small 82.6 & \small 82.6 & \small 80.38 & \small 76.4 & \small 79.6 & \small 79.5 & \small 76.9 & \small 78.8 & \small 79.4 & \small 72.73 & \small 73.2 & \small 79.2\\
\hline
\multirow{1}{*}{\small \bt-n} 
& - &  \small \textbf{83.7} &  \small \textbf{84.7} &  \small \textbf{83.4} &  \small \textbf{83.0} &  \small \textbf{82.7} &  \small \textbf{82.3} &  \small \textbf{80.6} &  \small \textbf{81.9} &  \small \textbf{82.9} &  \small \textbf{78.2} & \small \textbf{80.7} &  \small \textbf{83.4} &  \small \textbf{80.2} &  \small \textbf{81.6} & \small \textbf{82.1}\\
\hline
\multirow{1}{*}{\small \te-n} 
& - &  \small \textbf{83.8} & \small \textbf{84.8} & \small \textbf{83.5} &  \small \textbf{82.9} &  \small \textbf{83.7} &  \small \textbf{82.6} &  \small \textbf{81.2} &  \small \textbf{82.1} & \small \textbf{81.9} &  \small \textbf{79.2} & \small \textbf{81.3} &  \small \textbf{82.6} &  \small \textbf{78.1} &  \small \textbf{80.9} & \small \textbf{82.0}\\
\hline
\end{tabular}
\caption{\label{mthin}
Results of MT-hi-n (model trained on data translated to Hindi (en$\rightarrow$hi) using NLLB-translate) using different data settings \ref{sec:datasettings}.
}
\end{table*}

\begin{table*}
\centering
\begin{tabular}{|p{0.065\linewidth}|p{0.03\linewidth}p{0.03\linewidth}p{0.03\linewidth}p{0.03\linewidth}p{0.03\linewidth}p{0.03\linewidth}p{0.03\linewidth}p{0.03\linewidth}p{0.03\linewidth}p{0.03\linewidth}p{0.03\linewidth}p{0.03\linewidth}p{0.03\linewidth}p{0.03\linewidth}p{0.03\linewidth}p{0.03\linewidth}|}
\hline
\textbf{test} &  \textbf{en} & \textbf{fr} &  \textbf{es} & \textbf{de} & \textbf{el} & \textbf{bg} & \textbf{ru} & \textbf{tr} & \textbf{ar} & \textbf{vi} & \textbf{th} & \textbf{zh} & \textbf{hi} & \textbf{sw} & \textbf{ur} & \textbf{avg}\\
\hline
\multirow{1}{*}{\small \orig} 
&  \small 89.3 &  \small 83.5 &  \small 84.8 &  \small 83.4 &  \small 82.4 & \small 83.7 &  \small 80.5 & \small 79.4 &  \small 79.2 &  \small 79.9 & \small 78.3 & \small 79.4 & \small 77.2 & \small 72.7 & \small 74.0 & \small 80.5\\
\hline
\multirow{1}{*}{\small B-train} 
&  \small 89.2 &  \small 84.5 &  \small 85.9 &  \small 84.6 &  \small 84.3 &  \small \textbf{85.6} &  \small \textbf{82.9} &  \small 81.0 &  \small \textbf{81.8} &  \small \textbf{82.6} &  \small \textbf{79.8} &  \small \textbf{80.9} & \small \textbf{79.6} & \small 74.7 & \small 75.6 & \small \textbf{82.2}\\
\hline
\multirow{1}{*}{\small BT-enes} 
&  \small \textbf{89.8} &  \small \textbf{85.1} &  \small \textbf{86.2} &  \small 84.6 &  \small 84.1 &  \small 85.2 &  \small 82.4 &  \small \textbf{81.3} & \small 81.2 & \small 81.9 & \small 79.3 & \small \textbf{80.9} & \small 78.6 & \small \textbf{74.9} & \small \textbf{76.1} & \small 82.1\\
\hline
\multirow{1}{*}{\small \ttrain} 
&  \small 88.9 & \small 84.8 & \small 85.7 & \small \textbf{84.8} & \small \textbf{84.4} & \small 85.0 & \small 82.2 &  \small 80.9 &  \small 81.2 &  \small 81.9 & \small 78.9 & \small 80.7 &  \small \textbf{79.6} &  \small \textbf{74.9} & \small 75.9 & \small 81.9\\
\hline
\end{tabular}
\caption{\label{comparezs}
Comparing zero-shot test set results of different trained models (translations performed using NLLB).
}
\end{table*}

\begin{table*}
\centering
\begin{tabular}{| p{0.065\linewidth}|p{0.03\linewidth} p{0.03\linewidth}p{0.03\linewidth}p{0.03\linewidth}p{0.03\linewidth} p{0.03\linewidth}p{0.03\linewidth}p{0.03\linewidth}p{0.03\linewidth}p{0.03\linewidth}p{0.03\linewidth}p{0.03\linewidth}p{0.03\linewidth}p{0.03\linewidth}p{0.03\linewidth}p{0.03\linewidth}|}
\hline
\textbf{test} &  \textbf{en} & \textbf{fr} &  \textbf{es} & \textbf{de} & \textbf{el} & \textbf{bg} & \textbf{ru} & \textbf{tr} & \textbf{ar} & \textbf{vi} & \textbf{th} & \textbf{zh} & \textbf{hi} & \textbf{sw} & \textbf{ur} & \textbf{avg}\\
\hline
\multirow{1}{*}{\small \orig} 
& - & \small 82.1  &  \small 83.1 &  \small 80.7 & \small 82.3 &  \small 82.6 & \small 79.3  & \small 75.9 & \small 78.0 & \small 78.7 & \small 73.8 & \small 77.6 &  \small 77.7 & \small 70.5  & \small 71.3 & \small 78.1\\
\hline
\multirow{1}{*}{\small \btrain} 
& - &  \small 84.0 & \small \textbf{85.7} &  \small 82.4 &  \small 84.4 &  \small 84.4 &  \small 81.8 &  \small \textbf{78.9} &  \small 81.0 &  \small 80.9 & \small 77.4 & \small \textbf{80.5} &  \small 80.5 &  \small 73.6 & \small 74.4 & \small 80.7\\
\hline
\multirow{1}{*}{\small BT-enes} 
& - &  \small \textbf{84.2} &  \small 85.2 & \small \textbf{82.6} & \small \textbf{84.8} &  \small \textbf{84.8} & \small  \textbf{81.9} & \small  78.8 &  \small \textbf{81.7} &  \small \textbf{81.1} &  \small \textbf{78.2} &  \small 80.3 &  \small \textbf{80.7} &  \small \textbf{73.8} &  \small \textbf{75.1} & \small \textbf{80.9}\\
\hline
\multirow{1}{*}{\small \ttrain} 
& - &  \small 83.2 & \small 84.5 & \small 82.4 & \small 83.9 & \small 84.1 & \small 81.3 & \small 78.4 &  \small 80.6 & \small 80.7 & \small 76.6 & \small 79.7 & \small 80.1 & \small 73.1 & \small 74.2 & \small 80.2\\
\hline
\end{tabular}
\caption{\label{comparett}
Comparing translate-test (using NLLB translator) test set results of different trained models.
}
\end{table*}

\begin{table*}
\centering
\begin{tabular}{|p{0.065\linewidth}|p{0.03\linewidth}p{0.03\linewidth}p{0.03\linewidth}p{0.03\linewidth}p{0.03\linewidth}p{0.03\linewidth}p{0.03\linewidth}p{0.03\linewidth}p{0.03\linewidth}p{0.03\linewidth}p{0.03\linewidth}p{0.03\linewidth}p{0.03\linewidth}p{0.03\linewidth}p{0.03\linewidth}p{0.03\linewidth}|}
\hline
\textbf{test} &  \textbf{en} & \textbf{fr} &  \textbf{es} & \textbf{de} & \textbf{el} & \textbf{bg} & \textbf{ru} & \textbf{tr} & \textbf{ar} & \textbf{vi} & \textbf{th} & \textbf{zh} & \textbf{hi} & \textbf{sw} & \textbf{ur} & \textbf{avg}\\
\hline
\multirow{1}{*}{\small MT-hi-g} 
& \small \textbf{87.4} & \small 82.9 & \small \textbf{84.2} & \small 82.7 & \small \textbf{83.4} & \small \textbf{83.4} & \small 81.1 & \small \textbf{80.8} &  \small \textbf{79.9} &  \small \textbf{80.4} &  \small 78.1 & \small 79.9 &  \small \textbf{78.8} & \small \textbf{74.1} & \small \textbf{75.3} & \small \textbf{80.8}\\
\hline
\multirow{1}{*}{\small MT-hi-n} 
&  \small 87.2 & \small \textbf{83.4} & \small 83.6 & \small \textbf{82.9} & \small 82.7 & \small \textbf{83.4} & \small \textbf{81.8} & \small 79.9 & \small \textbf{79.9} & \small 80.1 &  \small \textbf{78.7} & \small \textbf{81.2} & \small 78.4 & \small 73.6 &  \small 74.9 & \small 80.7\\
\hline
\end{tabular}
\caption{\label{comparezshi}
Comparing zero-shot test set results of models trained on machine-translated Hindi~(1/3rd of training data), hi-g implies using google translator and hi-n implies using NLLB translator.
}
\end{table*}

\begin{table*}
\centering
\begin{tabular}{| p{0.065\linewidth}|p{0.03\linewidth} p{0.03\linewidth}p{0.03\linewidth}p{0.03\linewidth}p{0.03\linewidth} p{0.03\linewidth}p{0.03\linewidth}p{0.03\linewidth}p{0.03\linewidth}p{0.03\linewidth}p{0.03\linewidth}p{0.03\linewidth}p{0.03\linewidth}p{0.03\linewidth}p{0.03\linewidth}p{0.03\linewidth}|}
\hline
\textbf{test} &  \textbf{en} & \textbf{fr} &  \textbf{es} & \textbf{de} & \textbf{el} & \textbf{bg} & \textbf{ru} & \textbf{tr} & \textbf{ar} & \textbf{vi} & \textbf{th} & \textbf{zh} & \textbf{hi} & \textbf{sw} & \textbf{ur} & \textbf{avg}\\
\hline
\multirow{1}{*}{\small MT-hi-g} 
& - &  \small 81.7 & \small 82.6 & \small 80.1 & \small 82.2 & \small 82.3 & \small 80.3 & \small 76.2 &  \small 79.4 & \small 79.3 &  \small 77.9 &  \small 76.5 & \small 78.5 & \small 72.2 &  \small 72.5 & \small 78.7\\
\hline
\multirow{1}{*}{\small MT-hi-n} 
& - &  \small \textbf{82.2} & \small \textbf{83.6} & \small \textbf{80.6} & \small \textbf{82.6} & \small \textbf{82.6} & \small \textbf{80.4} & \small \textbf{76.4} &  \small \textbf{79.6} & \small \textbf{79.5} & \small \textbf{76.9} & \small \textbf{78.8} &  \small \textbf{79.4} & \small \textbf{72.7}  &  \small \textbf{73.2} & \small \textbf{79.2}\\
\hline
\end{tabular}
\caption{\label{comparetthi}
Comparing translate-test (using NLLB translator) test set results of models trained on machine-translated Hindi(1/3rd of training data), hig implies using google translator and hin implies using NLLB translator.
}
\end{table*}

\section{Comparing the Performance of mBert and XLMR \label{app:bert}}
As can be seen in Table \ref{mbertxlmr}, XLMR outperforms mBert by a huge margin on every language. Thus, we used XLMR for evaluating all our experiments.

\section{Performance of models using NLLB as the translator \label{app:nllb}}
Tables \ref{orig-nllb}, \ref{btrain-nllb} show the results of the models trained using \orig and \btrain training data.
Translation has been done using the NLLB translator. $\Delta$-n denotes the difference between max(\bt-n, \te-n) and max(\zs, \ttest). The results are similar to what we observe in Tables \ref{orig}, \ref{btrain}. $\Delta$-n is particularly high for low-resource languages like Hindi, Swahili, and Urdu. Also, the delta decreases for the \btrain model.

\section{Details of Model Training}
\label{appntraining}
The models mBert and XLMR were trained using the same setting as mentioned in the XTREME repository.%
\footnote{https://github.com/google-research/xtreme}%
 \paragraph{XNLI.} mBert is trained for 2 epochs with a learning rate of 2e-5, with a batch size of 8 and gradient accumulation of 4 (i.e an effective batch size of 32). XLMR is trained for 2 epochs with a learning rate of 5e-6, batch size of 5 and gradient accumulation steps of 6 (i.e effective batch size of 30). The final model is selected from the best checkpoint, which is based on the model's performance on the English dev set. For training the different variants of the model (\orig, \ttrain, \btrain, BT-enes, MT-hi-g, MT-hi-n) we use the same hyperparameter setting as mentioned above.\\
 We use xlm-roberta-large for all our experiments. Model training was done on a single Nvidia Geforce GTX 1080 Ti GPU, which has a RAM of 12GB. It took us around one day to train a single model for 2 epochs.
 For data translation using NLLB(3.3B parameter model), we made use of the NVIDIA A100-SXM4-80GB gpu for faster processing. Translating the test sets took couple of hours(1-1.5).
 \paragraph{MLQA.} To evaluate the performance on MLQA dataset, we trained XLMR on the SQUAD dataset~\citep{squad}. The model is trained for 3 epochs with a learning rate of 3e-5, batch size of 1 and gradient accumulation of 32~(i.e an effective batch size of 32).

 \paragraph{PAWS-X.} We trained xlm-roberta-large model on the English train set. The model is trained for 5 epochs with a learning rate of 2e-5, batch size of 2 and gradient accumulation of 16~(i.e an effective batch size of 32).

\section{Details of Train and Test Translations}
\label{appntranslation}
To train the model on back-translated (using Spanish as the pivot) and machine-translated(translated to Hindi and Spanish) data, we made use of the open-source 3.3B parameter NLLB model hosted on Hugging-Face%
\footnote{https://huggingface.co/facebook/nllb-200-3.3B}. 
 We found that the English to Chinese translation using NLLB is of lower quality, so we tried the open source 1.2B parameter M2M~\citep{fan2020englishcentric} 
 model
\footnote{https://huggingface.co/facebook/m2m100\_1.2B}%
 and it performed better compared to the NLLB translator.

\section{Details of Human Annotations}
\label{annotationdetails}
Each task (set of random 200 sentences) is annotated independently by two annotators. The task description shared with the annotators is included in Table~\ref{instructions}. The sentences in agreement between the two annotators are reviewed and approved for the dataset by the final annotator. If there is a mismatch, it is sent to the two annotators for review and possible corrections. If the mismatch persists, a third annotator performs a fresh annotation. The final annotator reviews the 3 answers and submits the final answer for the dataset. We also computed the Cohen's Kappa score between the two annotators and found them to be: 0.64 for English sentences, 0.43 for Hindi sentences, and 0.37 for Urdu sentences. Although the agreement scores are lower for Hindi and Urdu, for the machine-translated text they are still higher than human annotated text, especially for Urdu (0.41 for MT sentences vs. 0.37 for human translations). For the instances with conflicting labels from the two annotators, most of these instances were marked as neutral by one annotator and as entailment or contradiction by the other. A noticeable pattern for "neutral” versus “entailment" emerged: the hypothesis often included extra details or claims not explicitly stated in the premise. This tends to be labeled as neutral by the more meticulous annotator and as entailment when adopting a more flexible approach.

\section{Tools and Libraries}
We made use of awesome-align~\citep{dou2021word} to align words between English and any target language. The model used by awesome-align was bert-base-multilingual-cased. We used the Pytorch framework\footnote{https://pytorch.org/} and Hugging-face library\footnote{https://huggingface.co/} for all our model training and inferencing tasks. To integrate Labse~\citep{labse}, we made use of the Sentence-transformers library\footnote{https://www.sbert.net/}. To convert the transliterated sentences to the original scripts, we made use of both google-translate and Indic-trans \citep{Bhat:2014:ISS:2824864.2824872}~(for Indian languages). We made use of the google-cloud-translate api to use the google-translate services.


\section{More Trained Models}
\label{more_trained_models}
We trained a few more models in different settings to check their impact on the cross-lingual performance despite presence of semantic irregularities. The additional models we trained include:
\begin{enumerate}
    \item \ttrain is the model trained on English train set machine translated to Spanish. (See Table~\ref{ttrain}.)
    \item BT-enes, i.e train the model on backtranslated english (using Spanish as a pivot) + the original English.
    \item MT-hi-g, i.e train the model on machine-translated train set where the train set is translated to Hindi using google-translate. Here we used only 1/3rd of training data to train the model(to incur low costs of translation).
    \item MT-hi-n, this is the same as above, except that the translation is performed using NLLB translator.
\end{enumerate}
Using \ttrain is more effective in improving test performance across all target languages compared to using \orig \\
Tables \ref{btesen}, \ref{mthig}, \ref{mthin} shows the results of the trained models across different test settings~(test sets translated using NLLB). The figures highlight the potential semantic gap that exists between \bt and \ttest (also \zs and \te) across all the models which increases more towards the low resource languages. \\
In Table \ref{comparezs} and \ref{comparett}, we compare the zero shot and translate-test results of all the trained models across different languages. \btrain and BT-enes performs the best across majority of the languages. Table \ref{comparezshi}, \ref{comparetthi} compares the zero-shot and translate-test results of the MT-hi models, it can be seen that both the models perform equally across the languages, also because of training on less amount of data, their zero-shot performance is very slightly inferior to the \orig model. \\

\end{document}